%% file: egpaper_for_review.tex
\documentclass[10pt,twocolumn,letterpaper]{article}

\usepackage{iccv}
\usepackage{times}
\usepackage{epsfig}
\usepackage{graphicx}
\usepackage{amsmath}
\usepackage{amssymb}
\usepackage{booktabs}
\usepackage{makecell}
\usepackage{caption}
\input{math_commands.tex}

\usepackage{marvosym}
\usepackage[pagebackref=true,breaklinks=true,letterpaper=true,colorlinks,bookmarks=false]{hyperref}
\usepackage{multirow}

\usepackage[capitalize]{cleveref}
\crefname{section}{Sec.}{Secs.}
\Crefname{section}{Section}{Sections}
\Crefname{table}{Table}{Tables}
\crefname{table}{Tab.}{Tabs.}

\newenvironment{packed_itemize}{
 \vspace{-0.15cm}\begin{itemize}
  \setlength{\itemsep}{1pt}
  \setlength{\parskip}{0pt}
  \setlength{\parsep}{0pt}
 }{\end{itemize}}

\iccvfinalcopy

\begin{document}

\title{Text2Performer: Text-Driven Human Video Generation}

\author{Yuming Jiang$^{1}$
\quad
Shuai Yang$^{1}$
\quad
Tong Liang Koh$^{1}$
\quad
Wayne Wu$^{2}$
\quad
Chen Change Loy$^{1}$
\quad
Ziwei Liu\textsuperscript{1\Letter}\\
$^{1}$S-Lab, Nanyang Technological University
\quad
$^{2}$Shanghai AI Laboratory
\\
{\tt\small \{yuming002, shuai.yang, koht0029, ccloy, ziwei.liu\}@ntu.edu.sg \hspace{5pt} wuwenyan0503@gmail.com}\\
}

\twocolumn[{%
         \renewcommand\twocolumn[1][]{#1}%
         \maketitle
         \begin{center}
            \centering
            \vspace{-22pt}
            \includegraphics[width=0.99\textwidth]{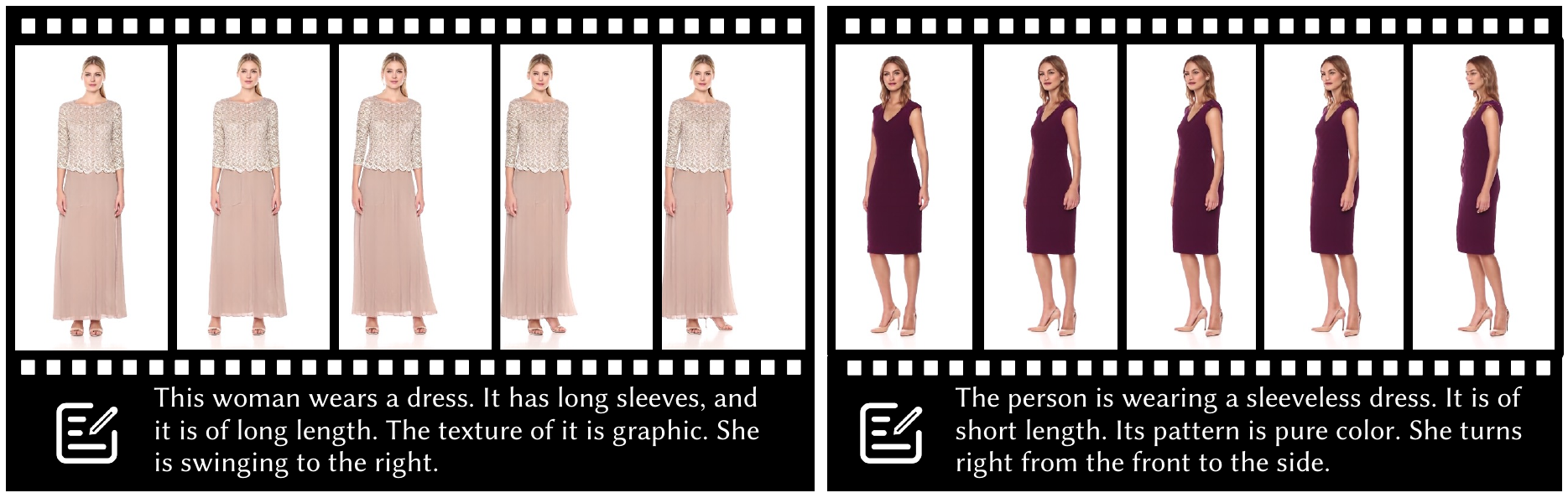}
            \vskip -0.25cm
            \captionof{figure}{\textbf{High-resolution videos generated by \textit{Text2Performer}}. The videos are generated by taking the text descriptions as the only input. The generated videos contain diverse appearances and flexible motions. Identities are well maintained.}
            \label{teaser}
         \end{center}%
      }]

\begin{abstract}
   Text-driven content creation has evolved to be a transformative technique that revolutionizes creativity.
   Here we study the task of text-driven human video generation, where a video sequence is synthesized from texts describing the appearance and motions of a target performer.
   Compared to general text-driven video generation, human-centric video generation requires maintaining the appearance of synthesized human while performing complex motions.
   In this work, we present \textbf{Text2Performer} to generate vivid human videos with articulated motions from texts.
   Text2Performer has two novel designs: \textbf{1)} decomposed human representation and \textbf{2)} diffusion-based motion sampler.
   First, we decompose the VQVAE latent space into human appearance and pose representation in an unsupervised manner by utilizing the nature of human videos.
   In this way, the appearance is well maintained
   along the generated frames.
   Then, we propose \textbf{continuous VQ-diffuser} to sample a sequence of pose embeddings.
   Unlike existing VQ-based methods that operate in the discrete space, continuous VQ-diffuser directly outputs the continuous pose embeddings for better motion modeling.
   Finally, motion-aware masking strategy is designed to mask the pose embeddings spatial-temporally to enhance the temporal coherence.
   Moreover, to facilitate the task of text-driven human video generation, we contribute a  Fashion-Text2Video dataset with manually annotated action labels and text descriptions.
   Extensive experiments demonstrate that Text2Performer generates high-quality human videos (up to $512\times256$ resolution) with diverse appearances and flexible motions. Our project page is \url{https://yumingj.github.io/projects/Text2Performer.html}.
\end{abstract}

\input{section/introduction}
\input{section/related_work}

\input{section/approach}
\input{section/dataset}

\input{section/experiment}

\input{section/conclusion}

{\small
   \bibliographystyle{ieee_fullname}
   \bibliography{egbib}
}
\newpage
\input{section/supp}

\end{document}

%% file: math_commands.tex
\usepackage{amsmath,amsfonts,bm}

\def\eqref#1{equation~\ref{#1}}

\def\1{\bm{1}}

\def\rmF{{\mathbf{F}}}

\def\rmI{{\mathbf{I}}}

\def\rmV{{\mathbf{V}}}

\def\ve{{\bm{e}}}
\def\vf{{\bm{f}}}

\def\vt{{\bm{t}}}

\DeclareMathAlphabet{\mathsfit}{\encodingdefault}{\sfdefault}{m}{sl}
\SetMathAlphabet{\mathsfit}{bold}{\encodingdefault}{\sfdefault}{bx}{n}



%% file: section/introduction.tex
\section{Introduction}
\label{sec:intro}

\begin{figure}
    \begin{center}
        \includegraphics[width=1.0\linewidth]{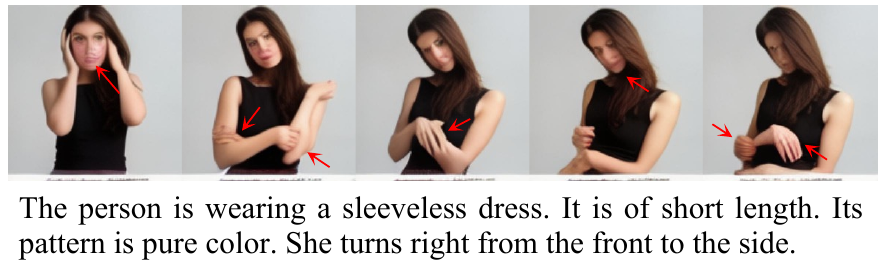}
    \end{center}
    \vspace{-17pt}
    \caption{\textbf{Results of General Large Text-to-Video Models.}
    We use the pretrained general large Text-to-Video Models~\cite{hong2022cogvideo} to generate videos using the same texts as the right example of Fig.~\ref{teaser}. The result fail to generate complete human structures and maintain temporal consistency.}
    \vspace{-15pt}
    \label{cogvideo}
\end{figure}

Since its emergence, text-guided image synthesis (\eg~DALLE~\cite{ramesh2021zero, ramesh2022hierarchical}) has attracted substantial attention. Recent works~\cite{saharia2022photorealistic, gafni2022make, ding2022cogview2, ding2021cogview, sun2022anyface, gu2022vector} have demonstrated fascinating performance for the quality of synthesized images and their consistency with the texts.
Beyond image generation, text-driven video generation~\cite{singer2022make, ho2022imagen, villegas2022phenaki, hong2022cogvideo} is an advanced topic to explore. 
Existing works rely on large-scale datasets to drive large models. Although they have achieved surprising performance on general objects,
when applied to the generation of some specific tasks, such as generating videos of garment presents for e-commerce websites, they fail to generate plausible results. Take the CogVideo~\cite{hong2022cogvideo} as an example. 
As shown in Fig.~\ref{cogvideo}, the generated human objects contain incomplete human structures, and the temporal consistency is poorly maintained.
On the other hand, these methods need billions of training data, which hampers the application to those specific tasks (\eg human video generation) without a large amount of paired data.

Therefore, it is worthwhile to explore text-to-video generation in human video generation, which has numerous applications~\cite{lee2019dancing,siyao2022bailando}.
In this paper, we focus on the task of text-driven human video generation.
Compared to general text-to-video generation, text-driven human video generation poses several unique challenges:
\textbf{1)} The human structure is articulated. The joint movements of different body components form many  complicated out-of-plane motions, \eg, rotations.
\textbf{2)} When performing complicated motions, the appearance of the synthesized human should remain the same.
For example, the appearance of a target human after turning around should be consistent with that at the first beginning.
In sum, to achieve high-fidelity human video generation, consistent human representation and complicated human motions should be carefully modeled.

We propose a novel text-driven human video generation framework \textbf{Text2Performer} to handle consistent human representations and complex human motions.
As shown in Fig.~\ref{teaser}, given texts describing appearance and motions, Text2Performer is able to generate temporally consistent human videos with complete human structures and unified human appearances.
Text2Performer is built upon VQVAE-based frameworks~\cite{bond2022unleashing, gu2022vector, esser2021taming, van2017neural}.
In Text2Performer, thanks to the specific nature of human videos which shares the same objects across the frames within one video, the VQVAE latent space can be decomposed into appearance and pose representations.
With the decomposed VQ-space, videos are generated by sampling appearance representation and a sequence of pose representations separately.
This decomposition contributes to the maintenance of human identity.
Besides, it makes the motion modeling more tractable, as the motion sampler does not need to take the appearance information into consideration.

To model complicated human motions, a novel continuous VQ-diffuser is proposed to sample a sequence of meaningful pose representations.
The architecture of the continuous VQ-diffuser is transformer.
The key difference to the previous transformer-based samplers~\cite{esser2021taming, van2017neural, bond2022unleashing} is that the continuous VQ-diffuser directly predicts the continuous pose embeddings rather than their indices in the codebook.
After predicting continuous pose embeddings,
we also make use of the rich embeddings stored in the codebook by retrieving the nearest embeddings of the predicted embeddings from the codebook.
Predicting continuous embeddings alleviates the one-to-many prediction issue in previous discrete methods and the use of codebook constrains the prediction space.
With this design, more temporally coherent human motions and more complete structures of human frames are sampled.
In addition, we borrow the idea of diffusion models~\cite{gu2022vector, bond2022unleashing, rombach2022high, ho2020denoising} to progressively predict the long sequence of the pose embeddings.
We propose a motion-aware masking strategy to sample the pose embeddings of the first frame and last frame firstly. Then the pose embeddings of the intermediate frames are gradually diffused.
The motion-aware masking strategy enhances the completeness of human structures and temporal coherence.

To facilitate the task of text-guided human video generation, we propose the Fashion-Text2Video Dataset.
It is built upon the FashionDataset~\cite{zablotskaia2019dwnet}, which consists of $600$ human videos performing the fashion show. We manually segment the whole video into clips and label the motion types. 
Each clip is performing one motion.
With the manually labeled motion labels, we then pair them with text descriptions.

Our contributions are summarized as follows:
\begin{packed_itemize}
    \item We present and study the task of text-guided human video generation. Our proposed \textbf{Text2Performer} can be well trained and have generative abilities with only a small amount of data for training.
    \item We propose to decompose the VQ-space into appearance and pose representations. The decomposition is achieved by making use of the nature of human videos, \ie, the motions are performed under one identity (appearance) across the frames. The decomposition of VQ-space improves the appearance coherence among frames and eases motion modeling.
    \item We propose the \textbf{continuous VQ-diffuser} to predict continuous pose embeddings with the pose codebook. The final continuous pose embeddings are iteratively predicted with the guidance of the motion-aware masking scheme. These designs contribute to generated frames of high quality and temporal coherence.
    \item We construct the Fashion-Text2Video Dataset with human motion labels and text descriptions to facilitate the research on text-driven human video generation.
\end{packed_itemize}

%% file: section/related_work.tex
\begin{figure*}
    \begin{center}
        \includegraphics[width=0.95\linewidth]{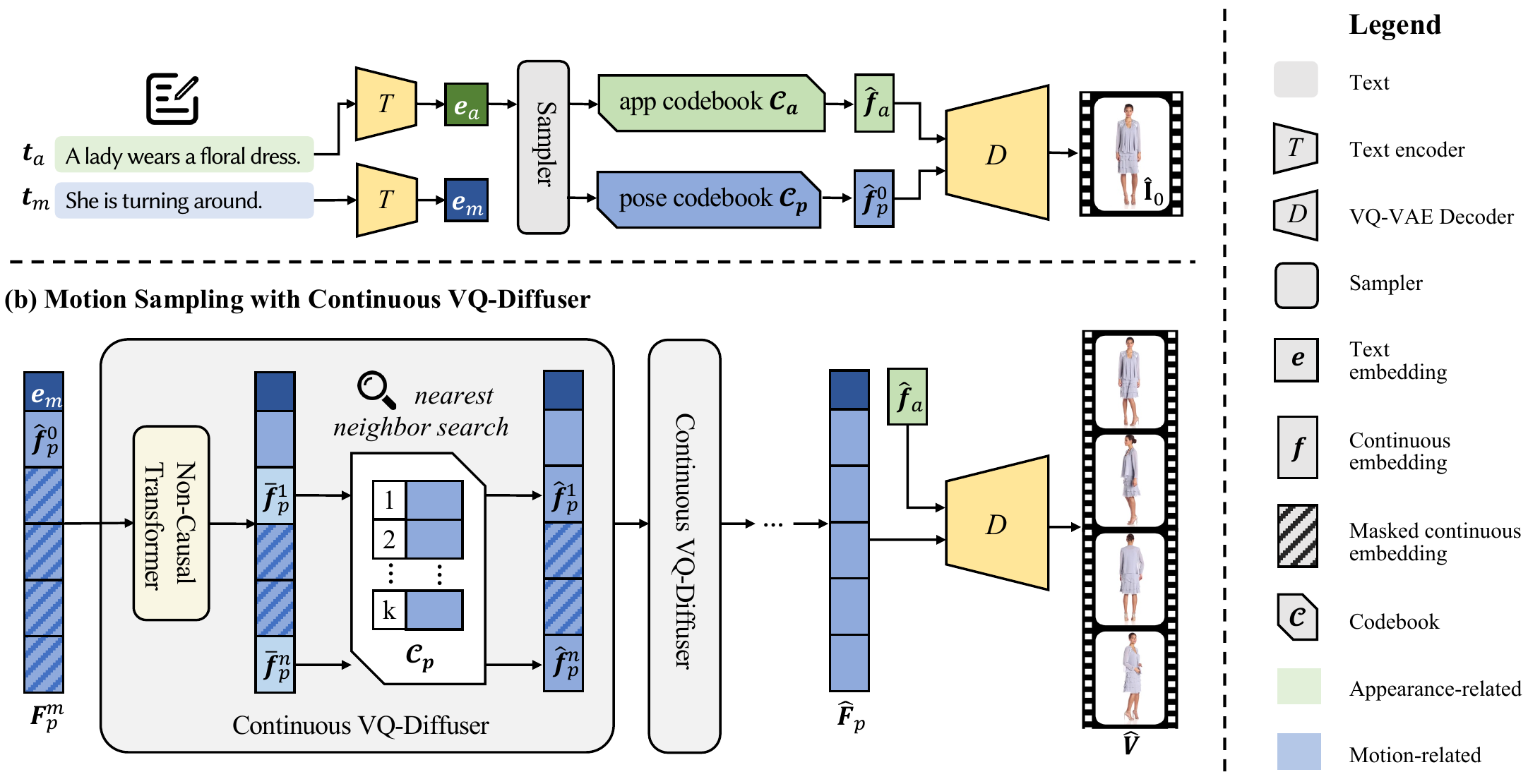}
    \end{center}
    \vspace{-20pt}
    \caption{\textbf{Overview of Text2Performer} with (a) Sampling from the Decomposed VQ-Space and (b) Motion Sampling with Continuous VQ-Diffuser.
        Given a text, we first sample the target appearance features $\hat{\vf_{a}}$ and exemplar pose features $\hat{\vf}^0_p$ %
        conditioned on the language features $\ve_a$ extracted by a pretrained text model.
        The motion sequence $\hat{\rmF}_p$ is then sampled by our proposed continuous VQ-Diffuser. The continuous VQ-Diffuser takes the extracted language features $\ve_m$ and $\hat{\vf}^0_p$ as inputs. The prediction of continuous motion sequences $\hat{\rmF}_p$ starts with fully masked pose features and  predicts continuous pose embeddings $\{\bar{\vf}^1_p, ..., \bar{\vf}^n_p \}$ with Non-Causal Transformer.
        The nearest neighbor pose embeddings $\{\hat{\vf}^1_p, ..., \hat{\vf}^n_p \}$ are then retrieved from the pose codebook $\mathcal{C}_p$.
        Guided by a motion-aware masking strategy, the continuous VQ-Diffuser is iteratively applied until the whole motion sequence is unmasked.
        The final videos are generated by feeding the continuous pose features and appearance features into the decoder of VQVAE.
    }
    \vspace{-10pt}
    \label{framework_illustration}
\end{figure*}

\section{Related Work}
\label{sec:related}

\noindent\textbf{Video Synthesis.}
Similar to image generation, GAN~\cite{goodfellow2020generative, brock2018large, karras2019style} and VQVAE~\cite{van2017neural, esser2021taming} are two common paradigms in the field of video synthesis~\cite{tulyakov2018mocogan}.
MoCoGAN-HD~\cite{tian2021good} and StyleGAN-V~\cite{skorokhodov2022stylegan} harness the powerful StyleGAN~\cite{karras2019style, karras2020analyzing} to generate videos.
DI-GAN~\cite{yu2022generating} models an implicit neural representation to generate videos. Brooks \etal \cite{brooks2022generating} propose to regularize the temporal consistency via a temporal discriminator.
These works focus on unconditional video generation.
As for the VQVAE-based methods, VideoGPT~\cite{yan2021videogpt} firstly extends the idea of VQVAE and autoregressive transformer to video generation. 
The following works introduce time-agnostic VQGAN with time-sensitive transformer~\cite{ge2022long} and masked sequence modeling~\cite{han2022show} to further improve the performance.
These VQVAE-based video synthesis frameworks accept conditions appended to the beginning of the token sequences.
Our proposed Text2Performer is developed on VQVAE. The differences lie in that the designs of VQ-space decomposition and continuous sampling.
Recently, there are some concurrent works~\cite{singer2022make, ho2022imagen, villegas2022phenaki, hong2022cogvideo} for text-to-video generation.
These methods are designed for general objects, while our Text2Performer has the design to separate human appearance and pose representations, which is specific to human video generation.

\noindent\textbf{Human Content Generation and Manipulation.}
Existing works focus on human images, and they are mainly divided into two types: unconditional generation and conditional generation. StyleGAN-Human~\cite{fu2022stylegan} employs StyleGAN to synthesize high-fidelity human images. TryOnGAN~\cite{lewis2021tryongan} and HumanGAN~\cite{sarkar2021humangan} generate human images conditioned on given human poses. Text2Human~\cite{jiang2022text2human} proposes to generate human images conditioned on texts and poses. We focus on human video generation.
For the human content manipulation, pose transfer~\cite{liu2019liquid, albahar2021pose, liu2020neural, liu2019neural, chan2019everybody} is a popular topic. Pose transfer deals with video data. The task is to transfer the poses from the reference videos to the source video.
Chan \etal \cite{chan2019everybody} take human poses as inputs and the desired video is generated by an image-to-image translation network.
Our task differs in that we only take texts as inputs and exemplar images are not compulsory.
Motion synthesis works~\cite{guo2020action2motion, petrovich2021action} generate human motions in form of 3D human representations (\eg, SMPL space and skeleton space), while our method generates human motions at the image level.

%% file: section/approach.tex
\section{Text2Performer}
\label{sec:approach}

\begin{figure}
    \begin{center}
        \includegraphics[width=1.0\linewidth]{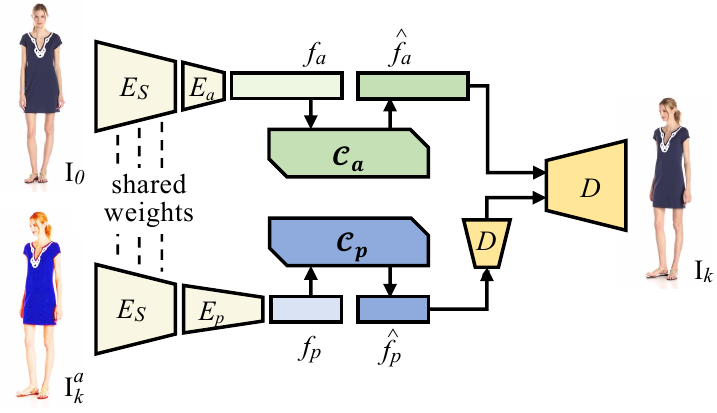}
    \end{center}
    \vspace{-20pt}
    \caption{\textbf{Pipeline of Decomposed VQ-Space.}
        Two different frames $I_0$ and $I_k$ of a video serve as the identity frame and the pose frame to provide appearance and pose features, respectively. We apply data augmentations to $I_k$ to erase its appearance information to avoid information leakage.
        Two codebooks are built to store the pose features and appearance features. The quantized features are finally fed into the decoder to reconstruct the pose frame $I_k$.}
    \vspace{-14pt}
    \label{decomposedVQ}
\end{figure}

As shown in Fig.~\ref{framework_illustration}, our proposed Text2Performer synthesizes desired human videos by taking the texts as inputs ($\vt_a$ for appearance and $\vt_m$ for motions).
To ensure consistent human representation, we propose to decompose the VQ-space into appearance representation $\vf_{a}$ and pose representation $\vf_{p}$ as shown in Fig.~\ref{decomposedVQ}.
With the decomposed VQ-space, we sample the human appearance features $\hat{\vf_{a}}$ and exemplar pose features $\hat{\vf}^0_p$ according to $\vt_{a}$.
To model the motion dynamics, we propose continuous VQ-diffuser as the motion sampler.
It progressively diffuses the masked sequence of pose features until the whole sequence is unmasked.
The final video clip is generated by feeding the appearance and pose embeddings into the decoder.

\subsection{Decomposed VQ-Space}

Human video generation requires consistent human appearance, \ie,
the face and clothing of the target person should remain unchanged while performing the motion.
Vanilla VQVAE~\cite{van2017neural} encodes images into a unified feature representation and builds a codebook on it.
However, such design is prone to generate drifted identities along the video frames.
To overcome this problem, we propose to decompose the VQ-space into the appearance representation $\vf_{a}$ and pose representation $\vf_{p}$ in an unsupervised manner. With the decomposed VQ-space, we can separately sample $\hat{\vf}_{a}$ and a sequence of $\hat{\vf}_{p}$ to generate the desired video.

In human video data, different frames in one video share the same human identity. We utilize this property of human video data to train the decomposed VQ-space.
The pipeline is shown in Fig.~\ref{decomposedVQ}. Given a video $\rmV = \{\rmI_0, \rmI_1, ..., \rmI_n\}$, we use its first frame $\rmI_{0}$ for appearance information and another randomly sampled frame $\rmI_k$ for pose information. 
$\rmI_{0}$ and $\rmI_{k}$ are fed into two encoders $E_{a}$ and $E_{p}$, respectively. The two branches share the encoder $E_s$.
To prevent $\rmI_{k}$ from leaking appearance information, data augmentations (\eg color jittering and Gaussian blur) are applied to $\rmI_{k}$ before the pose branch. $\vf_{a}$ and $\vf_{p}$ are obtained as
\begin{equation}
    \vf_a = E_a(E_s(\rmI_0)),
    \vf_p = E_p(E_s(\rmI^a_k))),
\end{equation}
where $\rmI^a_k$ is the augmented $\rmI_k$.
To make $\vf_p$ learn compact and necessary pose information, we make the spatial size of $\vf_p$ smaller. Given $\rmI \in \mathbb{R}^{H \times W \times 3}$, we empirically find that the optimal spatial sizes for $\vf_a$ and $\vf_p$ are $H / 16 \times W / 16$ and $H / 64 \times W / 64$, respectively.

After obtaining $\vf_a$ and $\vf_p$, two codebooks $\mathcal{C}_a$ and $\mathcal{C}_p$ are built to store the appearance and pose embeddings. The quantized feature $\hat{\vf}$ given codebook $\mathcal{C}$ is obtained by
\begin{equation}
    \hat{\vf} = Quant(\vf) := \{\underset{c_k \in \mathcal{C}}{\mathrm{argmin}} \left\| {f}_{ij} - c_k \right\|_2\ |  f_{ij} \in  \vf \}. 
\end{equation}
With the quantized $\hat{\vf}_a$ and $\hat{\vf}_p$, we then feed them into decoder $D$ to reconstruct the target pose frame $\hat{\rmI}_k$:
\begin{equation}
    \hat{\rmI}_k = D([\hat{\vf}_a, D_p(\hat{\vf}_p)]),
    \label{eq:rec}
\end{equation}
where $[\cdot]$ is the concatenation operation, and $D_p$ upsamples $\hat{\vf}_p$ to make it have the same resolution as $\hat{\vf}_a$.
The whole network (including the encoders, decoders, and codebooks) is trained using:
\begin{equation}
    \begin{split}
        \mathcal{L} = \left\| \rmI_k - \hat{\rmI}_k \right\|_1 &+ \left\| \text{sg}(\hat{\vf}_a) - \vf_a \right\|_2^2 + \left\| \text{sg}(\vf_a) - \hat{\vf}_a \right\|_2^2 \\ &+  \left\| \text{sg}(\hat{\vf}_p) - \vf_p \right\|_2^2 + \left\| \text{sg}(\vf_p) - \hat{\vf}_p \right\|_2^2, 
    \end{split}
\end{equation}
where $\text{sg}(\cdot)$ is the stop-gradient operation.

With the decomposed VQ-space, an additional sampler is trained to sample $\hat{\vf}_a$ and $\hat{\vf}^0_p$. This sampler has the same design as previous methods \cite{bond2022unleashing, jiang2022text2human}.

\vspace{-5pt}
\subsection{Continuous VQ-Diffuser}

We adopt the diffusion-based transformer~\cite{gu2022vector, bond2022unleashing} to sample the motion, a sequence of pose embeddings $\hat{\rmF}_p = \{\hat{\vf}^1_p, \hat{\vf}^2_p, ..., \hat{\vf}^n_p\}$ from the learned pose codebook $\mathcal{C}_p$. 
Different from autoregressive models~\cite{esser2021taming, van2017neural} that make predictions in a fixed order, diffusion-based transformer predicts multiple codebook indices in parallel. The prediction of codebook indices starts from $\rmF^0_p$, \ie, fully masked $\rmF^m_p$. The prediction at time step $t$ is represented as follows:
\begin{equation}
    \hat{\rmF}^t_p \sim q_\theta(\rmF^t_p | \rmF^{t-1}_p),
\end{equation}
where $\theta$ denotes the parameters of the transformer sampler.

To model human motions, we propose \textbf{1)} continuous space sampling, and \textbf{2)} motion-aware masking strategy.

\begin{figure}
    \vspace{-10pt}
    \begin{center}
        \includegraphics[width=0.97\linewidth]{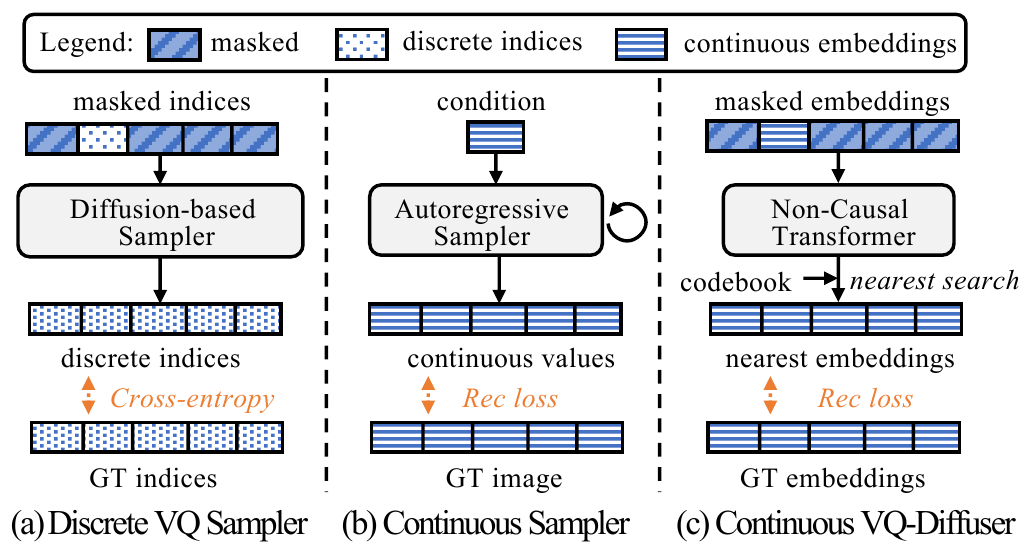}
    \end{center}
    \vspace{-20pt}
    \caption{\textbf{Comparison with Image Samplers.} (a) Discrete VQ Sampler predicts discrete codebook indices, which is trained by cross-entropy loss. (b) Continuous Sampler operates on unconstrained continuous RGB spaces. (c) Our VQ-Diffuser predicts continuous embeddings but also utilizes rich embeddings stored in codebooks.}
    \label{continuousVQ}
    \vspace{-12pt}
\end{figure}

\vspace{-6pt}
\subsubsection{Continuous Space Sampling}

As shown in Fig.~\ref{continuousVQ}(a), previous VQVAE-based methods~\cite{van2017neural, gu2022vector, bond2022unleashing, esser2021taming} sample tokens in a discrete space. They predict codebook indices to form quantized features, which are then fed into the decoder to generate images.
However, sampling at the discrete space is hard to converge because of redundant codebooks, which causes the one-to-many mapping prediction problem.
This would lead to meaningless predicted pose sequences.

In continuous VQ-diffuser $S_\theta$, we propose to train the non-causal transformer in the continuous embedding space as shown in Fig.~\ref{continuousVQ}(c). $S_\theta$ directly predicts continuous pose embeddings $\{\bar{\vf}^k_p\}$.
$S_\theta$ is trained to predict the full continuous $\bar{\rmF}_p$ from the masked pose sequences $\rmF^m_p$ conditioned on the motion description $\vt_m$ and the initial pose feature $\hat{\vf}^0_p$:
\begin{equation}
    \bar{\rmF}_p = S_\theta([T(\vt_m), \hat{\vf}^0_p, \rmF^m_p]),
\end{equation}
where $T(\cdot)$ is the pretrained text feature extractor. 

To utilize rich embeddings stored in $\mathcal{C}_p$ to constrain the prediction space, we retrieve the nearest embedding of the predicted continuous $\bar{\rmF}_p$ from $\mathcal{C}_p$ to obtain the final $\hat{\rmF}_p$:
\begin{equation}
    \hat{\rmF}_p = \text{Nearest}(\bar{\rmF}_p) := \{ \underset{c_k \in \mathcal{C}_p}{\mathrm{argmin}} \left\| \bar{\vf}_{p} - c_k \right\|_2 \}.
\end{equation}
$\hat{\rmF}_p$ is then fed into $D$ to reconstruct the final video $\hat{\rmV}$ following Eq.~(\ref{eq:rec}). Thanks to the continuous operation, we can add losses at both the image level and embedding level. The loss function to train our continuous VQ-diffuser is
\begin{equation}
    \begin{split}
        \mathcal{L} &= \left\| \bar{\rmF}_p - \rmF_p \right\|_1 + \left\| \text{sg}(\hat{\rmF}_p) - \bar{\rmF}_p \right\|_2^2 \\
        &+ \left\| \text{sg}(\bar{\rmF}_p) - \hat{\rmF}_p \right\|_2^2 + L_{\text{rec}}(\hat{\rmV}, \rmV),
    \end{split}
\end{equation}
where $L_{\text{rec}}$ is composed of $L_1$ loss and perceptual loss~\cite{johnson2016perceptual}. It should be noted that stop-gradient operation is applied to make the nearest operation differentiable.

Compared to continuous samplers like PixelCNN~\cite{van2016conditional} in Fig.~\ref{continuousVQ}(b), which directly predict the continuous RGB values, our continuous VQ-diffuser fully utilizes the rich contents stored in the codebook while restricting the prediction space. Therefore, it inherits the benefits of the discrete VQ sampler and continuous sampler at the same time.
The pose sequences sampled by our continuous VQ-diffuser are temporally coherent and meaningful.

\subsubsection{Motion-Aware Masking Strategy}

To generate plausible videos, the sampled motion sequence $\hat{\rmF}_p$ should be both temporally and spatially reasonable.
In order to make the continuous VQ-diffuser $S_\theta$ correctly condition on $\vt_m$ as well as generate reasonable human poses for each frame, we design a motion-aware masking strategy to meticulously mask $\rmF_p$ temporally and spatially.

At the temporal dimension, $S_\theta$ first predicts pose embeddings of the first frame and the last frame based on $\vt_m$ and $\hat{\vf}^0_p$. Then the prediction diffuses to the intermediate frames according to the given conditions and previously unmasked frames. Therefore, during the training, if the first frame and the last frame are masked, we will mask all frames to prevent the intermediate frames from providing information to help frame predictions at two ends.
A higher probability is assigned to masking all frames to help $S_\theta$ better learn prediction at the most challenging two ends.

At the spatial dimension, we propose to predict the first and the last frames with more than one diffusion step. Otherwise, the structure of the sampled human tends to be incomplete.
To this end, during the training, we mask the first and last frames partially at the spatial dimension.

The motion-aware masking strategy ensures the correctness of the motion in the temporal dimension and the completeness of the human structure in the spatial dimension.

%% file: section/dataset.tex
\section{Fashion-Text2Video Dataset}
\label{sec:dataset}

Existing human video datasets are either of low resolution~\cite{tulyakov2018mocogan, gorelick2007actions, balakrishnan2018synthesizing} or lack of text (action labels)~\cite{siarohin2019first, liu2019liquid}. To support the study of text-driven human video generation, we propose a new dataset, Fashion-Text2Video dataset.
Among existing human video datasets, we select videos in Fashion Dataset~\cite{zablotskaia2019dwnet} as the source data for following reasons: 
1) High resolution ($\geq1024 \times 1024$);
2) Diverse appearance.

The Fashion Dataset contains a total of $600$ videos. For each video, we manually annotate motion labels, clothing shapes (length of sleeves and length of clothing) 
and textures.
Motion labels are classified into 22 classes, including standing, moving right-side, moving left-side, turning around, \etc.
The lengths of sleeves are classified into no-sleeve, three-point, medium and long sleeves.
The length of clothing is the length of the upper part of the clothes.
Clothing textures are mainly divided into pure color, floral, graphic, \etc.
Upon the above labels, we then generate texts for each video using some pre-defined templates. The texts include descriptions of the appearances of the dressed clothes and descriptions for each sub-motions.
The proposed Fashion-Text2Video dataset can be applied to the research on Text2Video and Label2Video generation.

%% file: section/experiment.tex
\section{Experiments}
\label{sec:exp}

\begin{figure*}
    \vspace{-5pt}
   \begin{center}
      \includegraphics[width=0.90\linewidth]{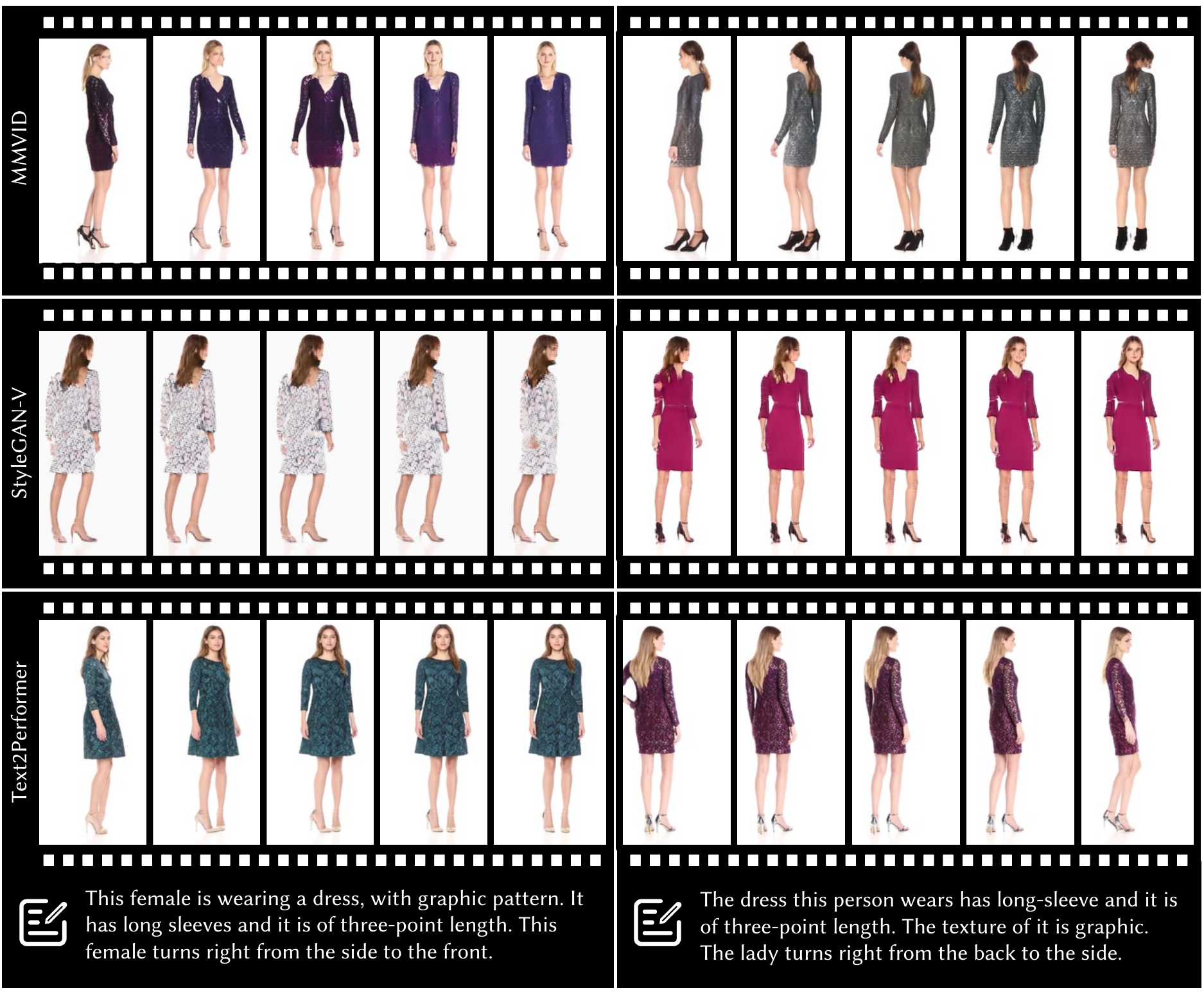}
   \end{center}
  \vspace{-20pt}
   \caption{\textbf{Qualitative Comparisons.}
   Text2Performer achieves superior generation qualities compared with baselines.
   }
  \vspace{-15pt}
   \label{visual_comp}
\end{figure*}

\subsection{Comparison Methods}
\textbf{MMVID}~\cite{han2022show} is a VQVAE-based method for multi-modal video synthesis. We follow the original setting to train MMVID on our dataset.
\textbf{StyleGAN-V}~\cite{skorokhodov2022stylegan} is a state-of-the-art video generation method built on StyleGAN. 
Action labels are used as conditions instead of texts, which we find significantly surpasses the performance of texts for this method.
\textbf{CogVideo}~\cite{hong2022cogvideo} is a large-scale pretrained model for Text-to-Video generation. 
We implement two versions of CogVideo, \ie, CogVideo-v1 and CogVideo-v2. 
Since the original CogVideo utilized the pretrained Text2Image dataset, we train CogVideo-v1 to use the exemplar image as an additional input to compensate for the lack of large-scale pretrained Text2Image models for human images. CogVideo-v2 is with original designs.

\subsection{Evaluation Metrics}
\noindent\textbf{Diversity and Quality.} 
\textbf{1)} We adopt Fréchet Inception Distance (FID)~\cite{heusel2017gans} to evaluate the quality of generated human frames. 
\textbf{2)} We evaluate the quality of generated videos using Fréchet Video Distance (FVD) and Kernel Video Distance (KVD)~\cite{unterthiner2018towards}. We generate $2,048$ video clips to compute the metrics, with $20$ frames for each clip.

\noindent\textbf{Identity Preservation.} We use two metrics to evaluate the maintenance of identity along the generated frames.
\textbf{1)} We extract face features of each generated frames using FaceNet~\cite{schroff2015facenet}. For each generated video clip, we calculate the $l_2$ distances between features of the frames and the first frame. 
\textbf{2)} We extract features of each generated frames using an off-the-shelf ReID model, \ie, OSNet~\cite{schroff2015facenet}. These features are related to identity features. We calculate the average cosine distances between extracted features of the frames and the first frame. 
The final metrics are obtained by averaging distances among $2,048$ generated video clips.

\noindent\textbf{User Study.} 
Users are asked to give three separate scores (the highest score is $4$) at three dimensions: 1) The consistency with the text for appearance, 2) The consistency with the text for desired motion, and 3) The overall quality of the generated video in terms of temporal coherence and its realism. 
A total of $20$ users participate in user study. Each user is presented with $30$ videos generated by different methods.

\begin{table}
  \centering
    \caption{\textbf{Quantitative Comparisons.}}
    \vspace{-8pt}
    \footnotesize{
    \begin{tabular}{l|c|c|c|c|c}
    \Xhline{1pt}
    \textbf{Method} & \textbf{FID $\downarrow$} & \textbf{FVD $\downarrow$} & \textbf{KVD $\downarrow$} &\textbf{Face $\downarrow$} & \textbf{ReID $\uparrow$} \\ \Xhline{1pt}
    MMVID~\cite{han2022show} & 11.85 & 303.02  & 78.67 & 0.9047 & 0.9096 \\ 
    StyleGAN-V~\cite{skorokhodov2022stylegan} & 29.68 & 219.63 & 18.77 & 0.8675 & \textbf{0.9568} \\ 
    CogVideo-v1~\cite{hong2022cogvideo} & 39.47 & 645.03 & 89.53 & 1.0564 & 0.8148 \\ 
    CogVideo-v2~\cite{hong2022cogvideo} & 51.76 & 799.80 & 112.24 & 1.0621 & 0.7960 \\ 
    \textbf{Text2Performer} & \textbf{9.60} & \textbf{124.78} & \textbf{17.96} & \textbf{0.8593} & 0.9382 \\
    \Xhline{1pt}
  \end{tabular}
  }
  \label{tab:quant_comp}
  \vspace{-5pt}
\end{table}

\begin{figure}
   \begin{center}
      \includegraphics[width=1.0\linewidth]{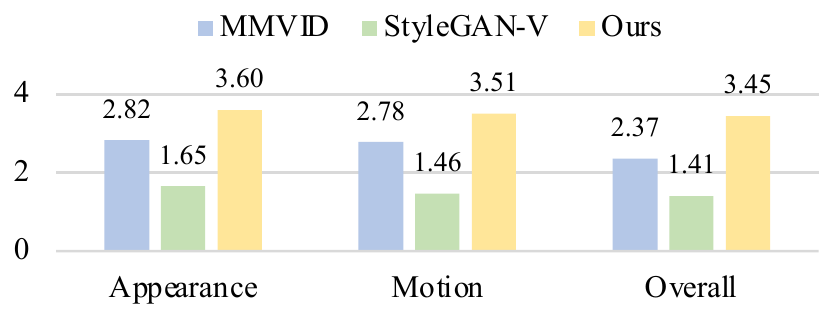}
   \end{center}
  \vspace{-20pt}
   \caption{\textbf{User Study.} 
   We achieve the highest scores.}
  \vspace{-15pt}
   \label{user_study}
\end{figure}

\subsection{Qualitative Comparisons}
We show two examples in Fig.~\ref{visual_comp}.
In the first example, MMVID and Text2Performer successfully generate desired human motions.
However, frames generated by MMVID have drifted identities. StyleGAN-V fails to generate plausible frames as well as consistent motions.
As for the second example, only Text2Performer generates corresponding motions. Artifacts appear in the video clips generated by MMVID and StyleGAN-V.
The visual comparisons demonstrate the superiority of Text2Performer.

\subsection{Quantitative Comparisons}
The quantitative comparisons are shown in Table~\ref{tab:quant_comp}.
In terms of FID, FVD and KVD, our proposed Text2Performer has significant improvements over baselines, which demonstrates the diversity and temporal coherence of the videos generated by our methods.
Existing baselines generate videos from entangled human representations and thus suffer from temporal incoherence.
As for the identity metrics, our method achieves the best performance on face distance and comparable performance on ReID distance with StyleGAN-V.
Since StyleGAN-V is prone to generate videos with small motions, features tend to be similar, leading to high ReID scores.
In our method, thanks to the decomposed VQ-space and continuous VQ-diffuser, temporal and identity consistency can be better achieved.
The result of the user study is shown in Fig.~\ref{user_study}.
Text2Performer achieves the highest scores on all three dimensions. Specifically, the result of the assessment on overall quality is consistent with the other quantitative metrics, which further validates the effectiveness of Text2Performer.

\vspace{-5pt}
\subsection{Ablation Study}

\begin{figure}
  \begin{center}
      \includegraphics[width=1.0\linewidth]{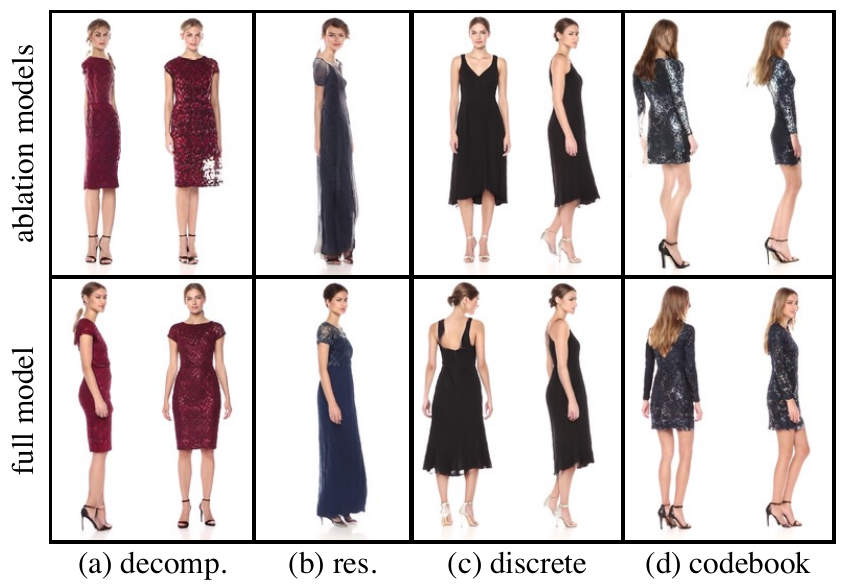}
  \end{center}
  \vspace{-20pt}
  \caption{\textbf{Ablation Studies} on (a) Decomposed VQ-Space, (b) Spatial Resolution of Pose Branch, (c) Discrete VQ sampler, and (d) Usage of Codebook.}
  \label{ablation_main}
  \vspace{-10pt}
\end{figure}

\noindent\textbf{Decomposed VQ-Space.}
The ablation model trains the continuous VQ-diffuser with a unified VQ-space, which fuses the identity and pose information within one feature. As shown in Fig.~\ref{ablation_main}(a), compared to our results, sampling in the unified VQ-Space generates drifted identity and incomplete body structures. This leads to inferior quantitative metrics in Table~\ref{tab:ablation}.
The unified VQ-Space requires the sampler to handle identity and pose at the same time, which poses burdens for the sampler.
With the decomposed VQ-Space, the identity can be easily maintained by fixing the appearance features, which further helps learn motions.

\noindent\textbf{Spatial Resolution of Pose Branch.}
In our design, the pose feature has $4\times$ smaller spatial resolution than the appearance feature. In Fig.~\ref{ablation_main}(b), we show the results generated by a decomposed VQVAE where pose and appearance features have the same resolution. Compared to results of the full model, 
body structures in the side view are incomplete. The quantitative metrics in Table~\ref{tab:ablation} and Fig.~\ref{ablation_main}(b) demonstrate that the smaller pose feature is necessary as the larger feature contains redundant information to harm valid feature disentanglement, thus adding challenges to the sampler.

\begin{figure*}
   \vspace{-5pt}
   \begin{center}
      \includegraphics[width=0.90\linewidth]{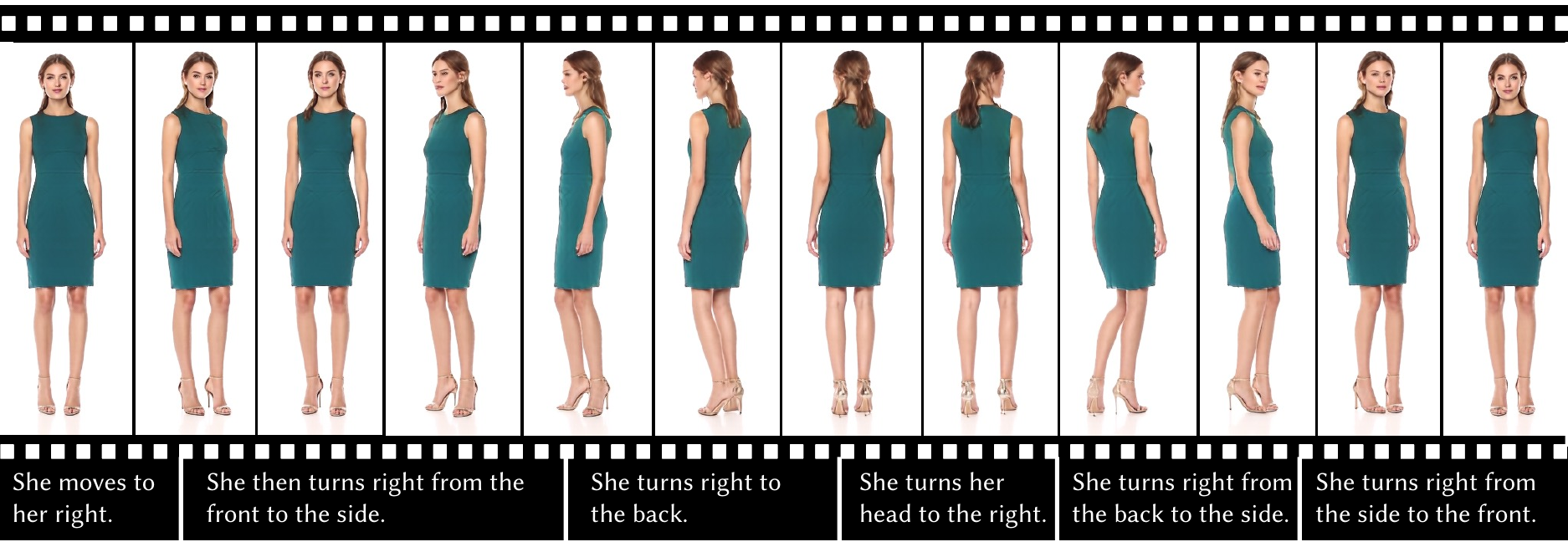}
   \end{center}
  \vspace{-17pt}
   \caption{\textbf{High-Resolution Results.}
   We show one result generated by a long text sequence.
   The identity is well maintained. 
   }
   \label{long_video}
  \vspace{-10pt}
\end{figure*}

\begin{figure*}
  \begin{center}
      \includegraphics[width=0.9\linewidth]{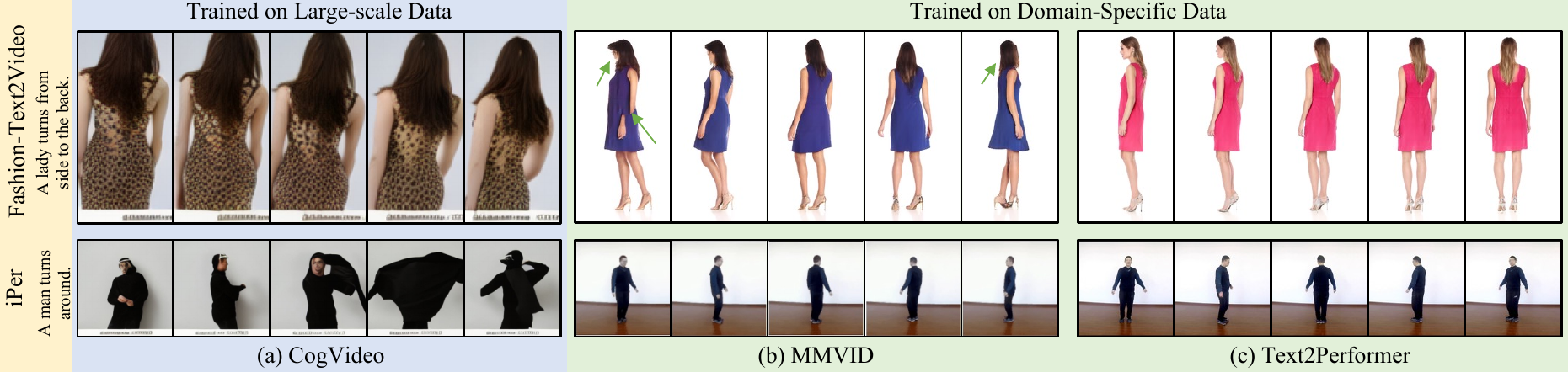}
  \end{center}
  \vspace{-18pt}
  \caption{\textbf{Cross-Dataset Performance.} We compare with models trained on large-scale data and domain-specific data.}
  \vspace{-15pt}
  \label{cross_data}
\end{figure*}

\begin{figure}
  \begin{center}
      \includegraphics[width=0.97\linewidth]{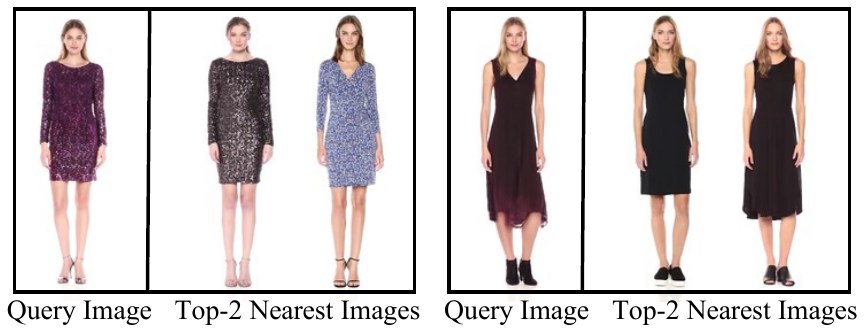}
  \end{center}
  \vspace{-18pt}
  \caption{\textbf{Nearest Neighbour Image Search.} Top-2 nearest images of generated images are retrieved. }
  \vspace{-18pt}
  \label{overfitting}
\end{figure}

\begin{table}
  \centering
    
    \caption{\textbf{Quantitative Results on Ablation Studies.}}
    \vspace{-10pt}
    \footnotesize{
    \begin{tabular}{l|c|c|c|c|c}
    \Xhline{1pt}
    \textbf{Method} & \textbf{FID $\downarrow$} & \textbf{FVD $\downarrow$} & \textbf{KVD $\downarrow$} & \textbf{Face $\downarrow$} & \textbf{ReID $\uparrow$} \\ \Xhline{1pt}
    w/o decomp. & 20.52 & 266.97  & 35.52 & 0.8936 & 0.9213 \\ 
    same res. & 15.69 & 202.06 & 32.96 &  0.8649  & 0.9399 \\
    disc. sampler & \textbf{10.37} & 308.88 & 73.17 & 0.9106  & 0.9161\\ 
    w/o codebook & 11.31 & 154.93 & 22.80 & 0.9437 & 0.9109 \\ 
    \textbf{Full Model} & 10.92 & \textbf{134.43} & \textbf{20.31} & \textbf{0.8520}  & \textbf{0.9421}\\
    \Xhline{1pt}
  \end{tabular}
  \label{tab:ablation}
  }
  \vspace{-15pt}
\end{table}

\noindent\textbf{Discrete VQ sampler in Decomposed VQ-Space.}
We train a variant of VQ-Diffuser to predict discrete codebook indices with the decomposed VQ-Space. In Fig.~\ref{ablation_main}(c), the discrete VQ Sampler can generate complete body structures but fails to generate video clips consistent with the given action control, \ie, turning right from the back to the side.
Failure on conditioning on action controls does not harm the quality of each generated frame, and thus this ablation study achieves a comparable FID score as our method. However, our method significantly outperforms this variant on other metrics as shown in Table~\ref{tab:ablation}.
In VQVAE, some codebook entries have similar contents but with different codebook indices. Therefore, treating the training of the sampler as a classification task makes the model hard to converge.

\noindent\textbf{Usage of Codebook in Continuous VQ-Space.}
In our design, after the prediction of embeddings, the nearest neighbor of embeddings will be retrieved from the codebook. 
We train an auto-encoder without codebook in the pose branch. The sampler is trained to sample continuous embeddings from the latent space of auto-encoder without the retrieval from codebook.
As shown in Fig.~\ref{ablation_main}(d), without the codebook, the sampler 
results in a disrupted human image at the side view. The quality of generated videos deteriorates as reflected in Table~\ref{tab:ablation}.
This demonstrates that predicting compact and meaningful continuous embeddings with codebooks enhances the quality of synthesized frames.

\subsection{Additional Analysis}
\noindent\textbf{High-Resolution Results.}
Text2Performer can generate long videos with high resolutions ($512 \times 256$) as shown in Fig.~\ref{teaser} and Fig.~\ref{long_video}.  Figure~\ref{long_video} shows an example of high-resolution videos for a long text sequence.

\noindent\textbf{Cross-Dataset Performance.}
We further verify the performance of models trained on different datasets. The results are presented in Fig.~\ref{cross_data}. Compared with CogVideo~\cite{hong2022cogvideo} pretrained on large-scale dataset, our Text2Performer and MMVID succeed to generate plausible specific human content. As for models trained on small-scale domain-specific data, our Text2Performer is comparable to MMVID on the simple iPer dataset~\cite{liu2019liquid}, and has superior performance on the more challenging Fashion-Text2Video dataset, especially on the completeness of body structures across frames. 

\noindent\textbf{Ability to Generate Novel Appearance.}
To verify that the generated humans are with novel appearances, we retrieve top-2 nearest neighbour images of generated frames using the perceptual distance~\cite{johnson2016perceptual}. As shown in Fig.~\ref{overfitting}, the generated query frames have different appearances from the top-2 nearest images retrieved from the dataset.

%% file: section/conclusion.tex
\vspace{-5pt}
\section{Discussions}
\label{sec:conclusion}

We propose a novel Text2Performer for text-driven human video generation.
We decompose the VQ-space into appearance and pose representations. Continuous VQ-diffuser is then introduced to sample motions conditioned on texts.
These designs empower Text2Perfomer to synthesize high-quality and high-resolution human videos.

\noindent\textbf{Limitations.}
On the one hand, Text2Performer is trained on videos with relatively clean backgrounds. 
On the other hand, the synthesized human videos are biased toward generating females with dresses. 
In future applications, more data can be involved in the training to alleviate the bias.

\noindent\textbf{Potential Negative Societal Impacts.}
The model may be applied to generate fake videos performing various actions. 
To alleviate the negative impacts, DeepFake detection methods can be applied to evaluate the realism of videos.

\noindent\textbf{Acknowledgement}.
This study is supported by the Ministry of Education, Singapore, under its MOE AcRF Tier 2 (MOE-T2EP20221-0012), NTU NAP, and under the RIE2020 Industry Alignment Fund - Industry Collaboration Projects (IAF-ICP) Funding Initiative, as well as cash and in-kind contribution from the industry partner(s). The study is also supported by the Google PhD Fellowship.

%% file: section/supp.tex
\onecolumn
\appendix
\section*{Supplementary}
\renewcommand\thesection{\Alph{section}}
\renewcommand\thefigure{A\arabic{figure}}
\renewcommand\thetable{A\arabic{table}}

In this supplementary file, we will explain our design of the proposed continuous VQ-diffsuer for handling interpolation and sampling within one model in Section \ref{interpolation}. 
Then we will introduce implementation details in Section \ref{implementation}. 
In Section \ref{experiment}, we will give more detailed explanations on evaluation metrics. 
In Section \ref{masking}, we provide more analysis on motion-aware masking strategy.
In Section \ref{sec:more_comp}, we provide more qualitative and quantitative comparisons with baseline methods.
Then we provide more visual results in Section \ref{results}. 
In Section \ref{limitation}, we will discuss the limitations of this work. 

\section{Interpolation and Sampling within One Model in Continuous Diffuser}
\label{interpolation}

During the training of the continuous VQ-diffuser $S_\theta$, video clips are normalized to a fixed number of video frames. The normalization guarantees that the first frame and the last frame are the exact initial and ending poses corresponding to the given motion text $\vt_m$.
However, videos generated in this way are not temporally smooth since the normalization operation extracts unconsecutive frames from the original video clips.

To further improve the temporal consistency and support scalable frame rates, we train $S_\theta$ to perform video interpolation as well.
Specifically, for the interpolation task, we set the text description $\vt_m$ as a predefined text ``empty'' and randomly mask the pose embeddings of intermediate frames. In this mode, the pose embeddings at the two ends (\ie, the first frame and the last frame) are always unmasked as the predictions of the intermediate frames are supposed to be conditioned on the neighbouring unmasked frames.
The interpolation task is trained at a lower probability than the sampling task since the interpolation task is simpler compared to the generation task.

The training video clips are mixed with normalized video clips and original video clips.
When normalized video clips are fed into the framework, the corresponding text descriptions are $\vt_m$.
If the original video clips are provided, the text description $\vt_m$ will be replaced with ``empty'' in our setting.

\section{Implementation Details}
\label{implementation}
Our main experiments are conducted on our proposed Fashion-Text2Video dataset. 
The Fashion-Text2Video dataset contains $600$ videos in total.
We exclude 5 videos from training. The excluded videos are used for visualizing the performance of trained decomposed VQVAE and VQ-Sampler on unseen data.
Models are trained with $4$ Tesla V100 GPUs.
Our main experiments are conducted on videos with resolution of $256\times128$.
The decomposed VQVAE is trained with batch size of $64$ for $145, 000$ iterations.
The sampler for exemplar pose and appearance is trained with batch size of $4$. The texts for each exemplar image are generated on-the-fly for generalizability.
The continuous VQ-Diffuser is trained with batch size of $4$.
The probability of training continuous VQ-Diffuser with interpolation mode is set to $0.2$. 
The probability of masking all temporal frames is set to $0.375$. The diffusion steps for sampling the frames at the two ends are set to $6$ at the spatial dimension.

\section{Further Explanations on Evaluation Metrics}
\label{experiment}

\noindent\textbf{Diversity and Quality.} As described in Section {\color{red}5.2} of our main paper, we adopt FID~\cite{heusel2017gans}, FVD and KVD~\cite{unterthiner2018towards} as the metrics to indicate the diversity and quality of synthesized videos. We extract $2, 048$ video clips from the original dataset as the real videos. Each video clip has $20$ frames. We then use the text descriptions corresponding to the extracted $2, 048$ video clips to synthesize videos.
To calculate FID, we use the all frames as the images, \ie, we calculate the FID values on $40, 960$ images.
For FVD and KVD, we directly compute the values on $2,048$ video clips.

\noindent\textbf{Identity Preservation.}
For face scores~\cite{schroff2015facenet}, we first detect the faces and then extract the features for the cropped faces. Since there are frames without detectable faces (the target person rotates to the back view), we compute the distances between the frames with detectable faces and the first frame with detectable faces in one video clip. We then use the average of these distances as the final score. 
For ReID scores, features are extracted on the original generated frames~\cite{schroff2015facenet}, and the scores are computed on the extracted features.

\section{Further Analysis on Motion-Aware Masking Strategy}
\label{masking}

\begin{figure}
  \begin{center}
      \includegraphics[width=0.5\linewidth]{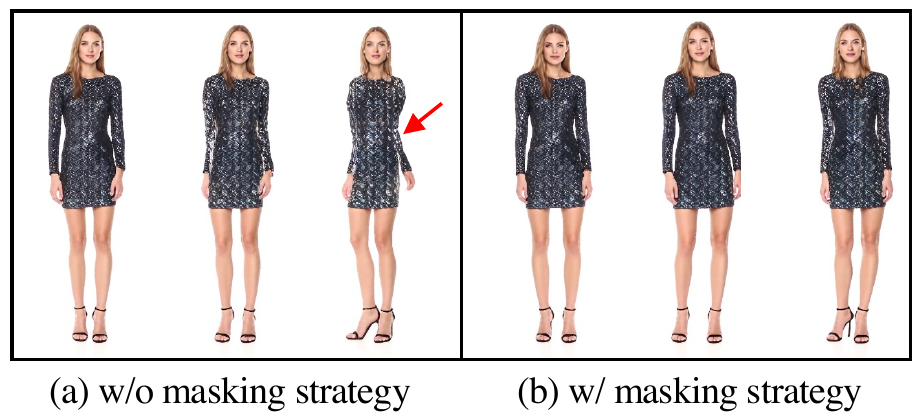}
  \end{center}
  \vspace{-15pt}
  \caption{\textbf{Ablation Studies on Motion-aware Masking Strategy.}}
  \label{ablation_masking}
\end{figure}

We conduct further analysis on the motion-aware masking strategy. As shown in Table~\ref{tab:masking}, after adopting the motion-aware masking strategy, the performance improves slightly. Using motion-aware masking strategy achieves superior FID, KVD, Face scores and ReID scores. 
We further demonstrate the effectiveness of motion-aware masking strategy in Fig.~\ref{ablation_masking} as the improvements cannot be well reflected in the quantitative results. As shown in Fig.~\ref{ablation_masking}, the use of motion-aware masking strategy enhances of the completeness of the body structures.

\begin{table*}
  \centering
    \caption{\textbf{Further Analysis on Masking Strategy}}
    \vspace{-10pt}
    \begin{tabular}{l|c|c|c|c|c}
    \Xhline{1pt}
    \textbf{Method} & \textbf{FID $\downarrow$} & \textbf{FVD $\downarrow$} & \textbf{KVD $\downarrow$} & \textbf{Face $\downarrow$} & \textbf{ReID $\uparrow$} \\ \Xhline{1pt}
    w/o masking & 14.31 & \textbf{122.93} & 22.50 & 0.7617 & 0.9494 \\ 
    \textbf{Full Model} & \textbf{14.19} & 125.38 & \textbf{22.27} & \textbf{0.7562}  & \textbf{0.9522} \\
    \Xhline{1pt}
  \end{tabular}
  
  \label{tab:masking}
\end{table*}

\section{More Comparisons}
\label{sec:more_comp}

\begin{figure}[t!]
  \begin{center}
      \includegraphics[width=0.85\linewidth]{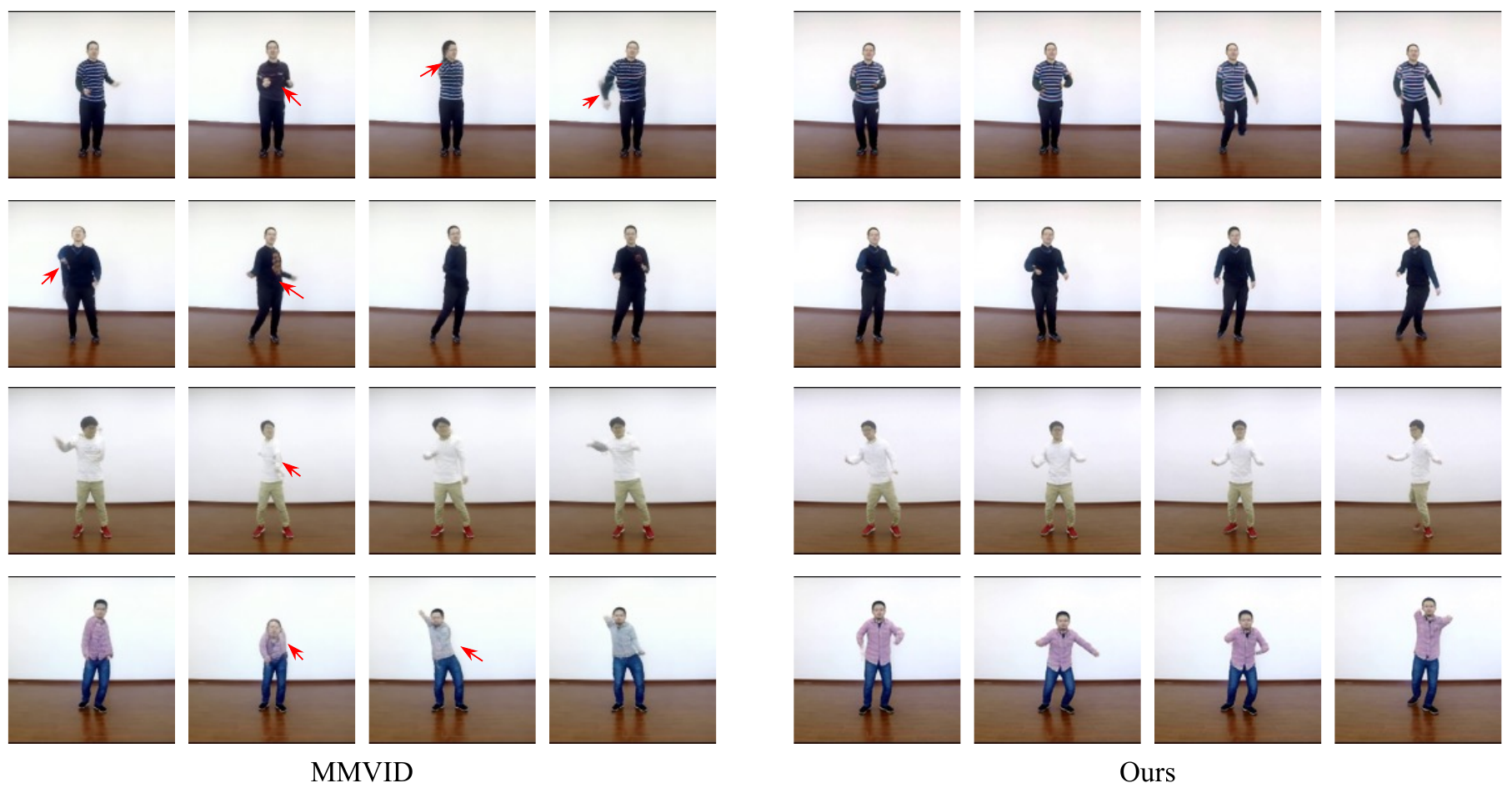}
  \end{center}
  \vspace{-15pt}
  \caption{\textbf{Qualitative Comparisons with MMVID~\cite{han2022show} on iPer Dataset~\cite{liu2019liquid}.}}
  \label{comp_iper}
\end{figure}

In Table~\ref{tab:more_quant_comp}, we present more quantitative comparisons. Apart from our proposed Fashion Text2Video dataset, we also compare with MMVID~\cite{han2022show} on iPer dataset~\cite{liu2019liquid}. We use the original implementation of MMVID and retrain the method as no pretrained model is provided for this dataset. As shown in Table~\ref{tab:more_quant_comp}, our method achieves a better FID score as we synthesize frames with higher quality. We also achieve a comparable FVD score with MMVID. We attribute the value gap of FVD to the nature of FVD metrics. We find that the FVD score focuses more on the motion than the appearance consistency on this dataset. Since the resolution is relatively low on this dataset, some appearance flickering artifacts generated by MMVID are deemed as motions, and thus result in a slightly lower FVD score. Qualitative comparisons can be found in Fig.~\ref{comp_iper}.

\section{More Qualitative Results}
\label{results}
\begin{figure}[h!]
  \begin{center}
      \includegraphics[width=0.85\linewidth]{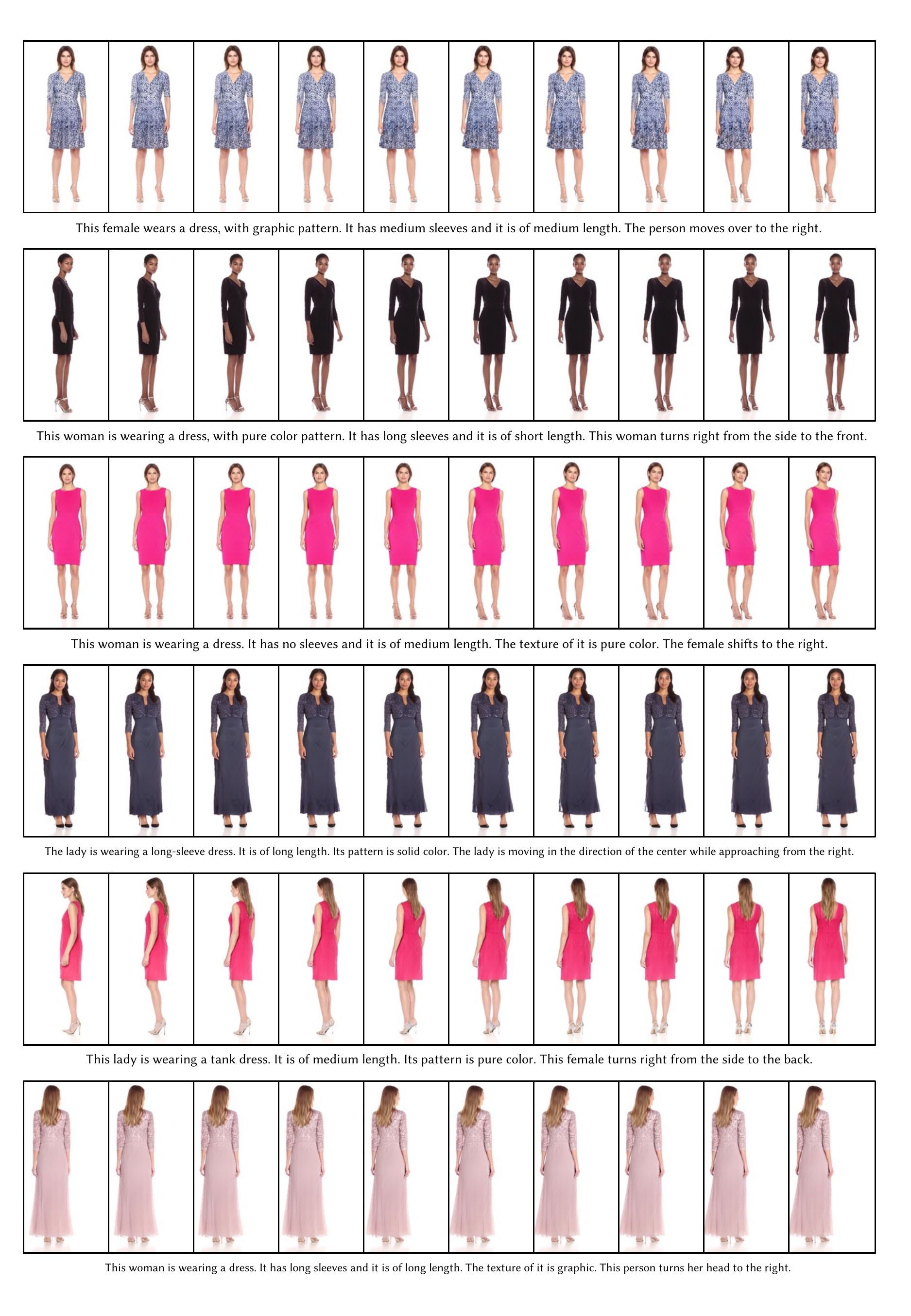}
  \end{center}
  \vspace{-15pt}
  \caption{\textbf{Our generated videos with the resolution of $256 \times 128$.}}
  \label{ours_256}
\end{figure}

\begin{figure}[h!]
  \begin{center}
      \includegraphics[width=1.0\linewidth]{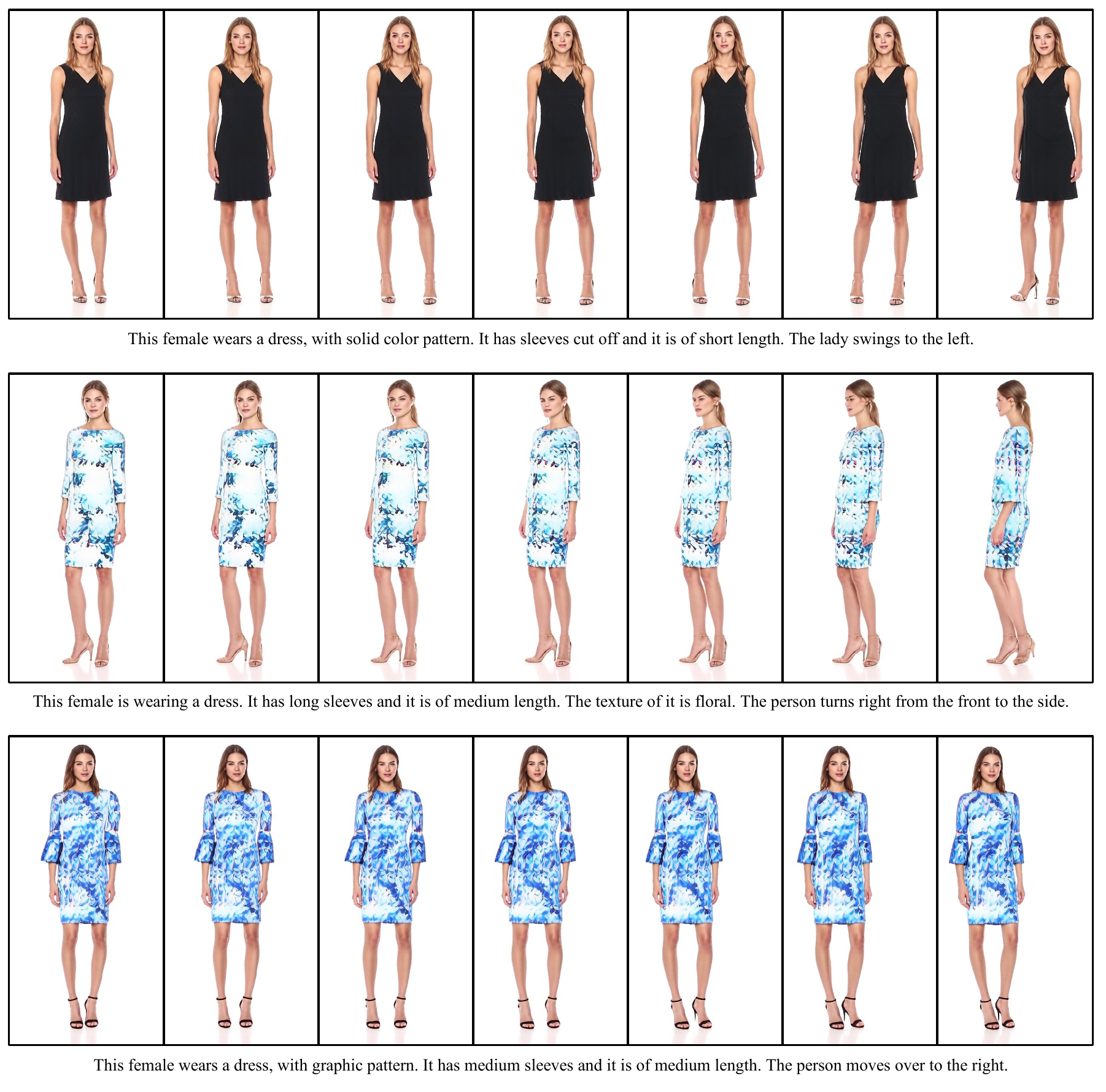}
  \end{center}
  \vspace{-15pt}
  \caption{\textbf{Our generated videos with the resolution of $512 \times 256$.}}
  \label{ours_512}
\end{figure}

\begin{figure}[h!]
  \begin{center}
      \includegraphics[width=1.0\linewidth]{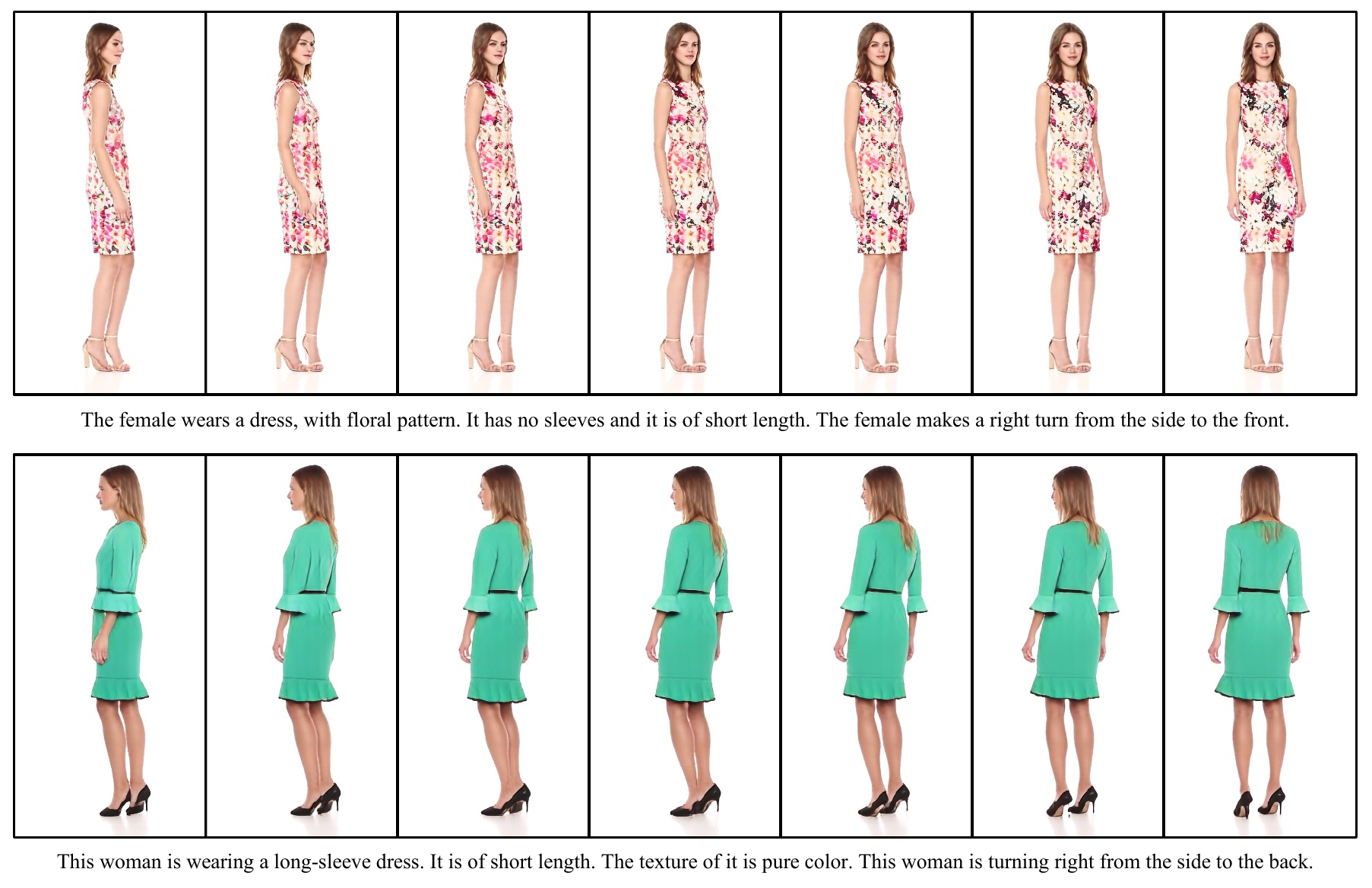}
  \end{center}
  \vspace{-15pt}
  \caption{\textbf{Our generated videos with the resolution of $512 \times 256$.}}
  \label{ours_512_2}
\end{figure}

\begin{figure}[h!]
  \begin{center}
      \includegraphics[width=0.85\linewidth]{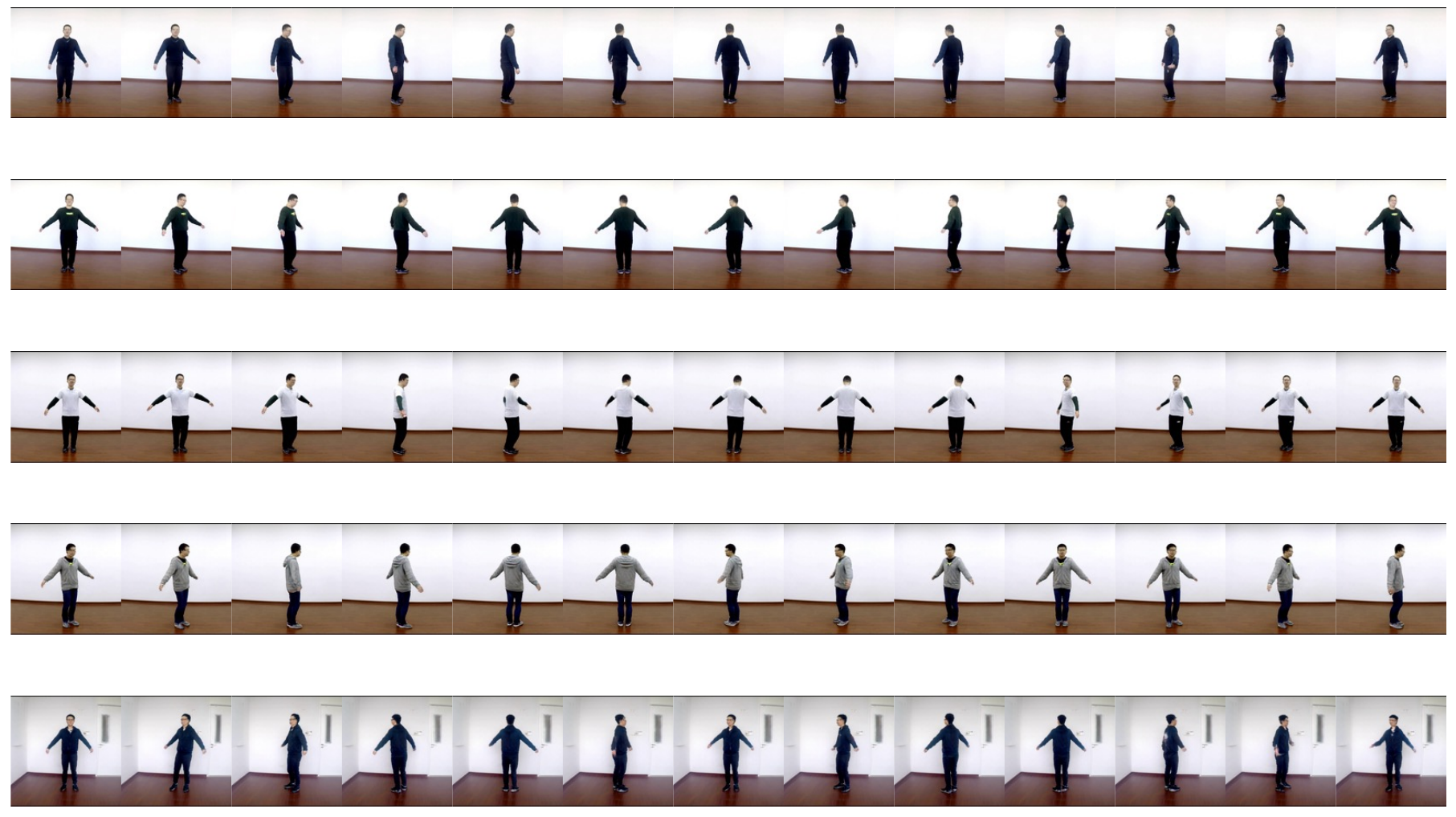}
  \end{center}
  \vspace{-15pt}
  \caption{\textbf{Our results on iPer Dataset~\cite{liu2019liquid}.}}
  \label{ours_iper}
\end{figure}

We show more results in Figs.~\ref{ours_256}-\ref{ours_512_2}.
Figure~\ref{ours_256} shows the video clips with resolution of $256 \times 128$. Video clips in Figs.~\ref{ours_512}-\ref{ours_512_2} are with the resolution of $512 \times 256$.
We also show the results on the iPerDataset in Fig.~\ref{ours_iper}, where we follow the same setting as MMVID~\cite{han2022show} to train and generate videos.

\begin{table*}
  \centering
    \caption{\textbf{More Quantitative Comparisons.}}
    \vspace{-8pt}
    \begin{tabular}{c|l|c|c}
    \Xhline{1pt}
    \textbf{Dataset} & \textbf{Method} & \textbf{FID $\downarrow$} & \textbf{FVD $\downarrow$}  \\ \Xhline{1pt}
    \multirow{5}{*}{Fashion-Text2Video} & StyleGAN-V~\cite{skorokhodov2022stylegan} & 29.68 & 219.63   \\ 
    & CogVideo-v1~\cite{hong2022cogvideo} & 39.47 & 645.03  \\ 
    & CogVideo-v2~\cite{hong2022cogvideo} & 51.76 & 799.80   \\ 
    & MMVID~\cite{han2022show}  & 11.85  & 303.02    \\ 
    & \textbf{Text2Performer} & \textbf{9.60} & \textbf{124.78}   \\ \hline
    \multirow{2}{*}{iPer~\cite{liu2019liquid}} & MMVID~\cite{han2022show} & 30.80 & \textbf{256.72}  \\ 
    & \textbf{Text2Performer} & \textbf{15.47}  & 289.69 \\ \hline
    \Xhline{1pt}
  \end{tabular}

  \label{tab:more_quant_comp}
  \vspace{-5pt}
\end{table*}

\section{Limitations}
\label{limitation}
On the one hand, Text2Performer is trained on videos with relatively clean background. In future works, more designs should be introduced to handle the complex background.
On the other hand, the synthesized human videos are biased toward generating females with dresses. This is because the original FashionDataset~\cite{zablotskaia2019dwnet} only contains videos of females with dresses. 
In future works, more data can be involved in the training to alleviate the issue caused by the dataset bias.

